\documentclass[preprint,12pt]{elsarticle}

\usepackage{amssymb}
\usepackage{amsmath}
\usepackage{algorithmic}
\usepackage{algorithm}
\usepackage{url}
\usepackage{booktabs}
\usepackage{subfigure}
\usepackage{bbding}
\usepackage{pifont}
\usepackage{multirow}

\journal{xxx}

\begin{document}

\begin{frontmatter}



\title{Mutual Information-based Representations Disentanglement for Unaligned Multimodal Language Sequences}


\author{Fan Qian, Jiqing Han, Jianchen Li, Yongjun He, Tieran Zheng, Guibin Zheng} 

\affiliation{organization={Harbin Institute of Technology},
            city={Harbin},
            postcode={150001}, 
            country={China}}

\begin{abstract}
The key challenge in unaligned multimodal language sequences lies in effectively integrating information from various modalities to obtain a refined multimodal joint representation. 
Recently, the \textit{disentangle and fuse} methods have achieved the promising performance by explicitly learning modality-agnostic and modality-specific representations and then fusing them into a multimodal joint representation.
However, these methods often independently learn modality-agnostic representations for each modality and utilize orthogonal constraints to reduce linear correlations between modality-agnostic and modality-specific representations, neglecting to eliminate their nonlinear correlations. 
As a result, the obtained multimodal joint representation usually suffers from information redundancy, leading to overfitting and poor generalization of the models. 
In this paper, we propose a Mutual Information-based Representations Disentanglement (MIRD) method for unaligned multimodal language sequences, in which a novel disentanglement framework is designed to jointly learn a single modality-agnostic representation.
In addition, the mutual information minimization constraint is employed to ensure superior disentanglement of representations, thereby eliminating information redundancy within the multimodal joint representation.
Furthermore, the challenge of estimating mutual information caused by the limited labeled data is mitigated by introducing unlabeled data.
Meanwhile, the unlabeled data also help to characterize the underlying structure of multimodal data, consequently further preventing overfitting and enhancing the performance of the models.
Experimental results on several widely used benchmark datasets validate the effectiveness of our proposed approach.
\end{abstract}


\end{frontmatter}



\section{Introduction}
Human communication occurs through not only spoken language but also non-verbal behaviors such as facial expressions and vocal intonations \cite{Milo1995ToolsLA}. 
This form of language is referred to as multimodal language in computational linguistics \cite{MOSEI,RMFN}. 
With the increasing amount of user-generated multimedia content, Multimodal Sentiment Analysis (MSA) is emerging as a popular research field and one of the most important application directions in multimodal language \cite{FMSA}.
MSA aims to understand the sentimental tendencies conveyed in opinion videos by leveraging information from multiple modalities, including visual, language and audio.
It has broad applications in various fields, such as e-commerce, public opinion monitoring, human-computer interaction, etc.

Early works \cite{Chen2017MultimodalSA,MFN,MARN,MCTN,MFM,RAVEN,TMA} in MSA primarily focused on aligned multimodal language sequences, which are formed by manually aligning visual and audio features to the word granularity based on the time interval of each word, typically by taking the average of visual and audio features. 
However, this manual alignment process requires domain-specific knowledge and consumes significant time and effort. 
Consequently, current research has gradually shifted its focus to unaligned multimodal language sequences, where the key challenge lies in effectively integrating information from various modalities (i.e., multimodal fusion) to obtain a refined multimodal joint representation \cite{Self-MM,MMIM,MTSA}.

Previous fusion methods can be regarded as a form of \textit{directly fuse},  which refers to the fusion of information from different modalities without preprocessing.
Based on different learning strategies, the \textit{directly fuse} methods for unaligned multimodal language sequences can be roughly categorized into three types: tensor-based methods, graph-based methods, and cross-modal attention-based methods.
Tensor-based methods \cite{TFN,LMF} utilize the tensor product of the representations encoded from each modality to capture cross-modal interactions, resulting in a high-dimensional multimodal joint representation
Graph-based methods either construct a separate graph for each modality, modeling the interactions within each modality and then aggregating the graph representations of each modality to obtain a multimodal joint representation \cite{GPFN,GraphCAGE}, or construct a single graph for sequential elements of all modalities, aiming to model the interactions across modalities and learn a refined graph representation \cite{MTGAT}.
The core of cross-modal attention-based methods \cite{MulT,PMR,LMR-CBT} is to model long-term dependencies across modalities. 
By repeatedly enhancing the features of one modality with the features of another modality, the enhanced sequential representations can be obtained for each modality, which are then aggregated into a multimodal joint representation. 
Although these \textit{directly fuse} methods have achieved a good performance, they overlook the interference caused by modality heterogeneity and distribution gap. 
That is to say, they do not explicitly consider the shared factor between modalities and have poor interpretability.

Therefore, the \textit{disentangle and fuse} methods have recently begun to emerge, with the aim of simultaneously exploring the commonality and complementarity of different modalities before multimodal fusion.
For example, in terms of commonality, it includes the motivations and intentions of the speaker \cite{MISA}. 
Regarding complementarity, it encompasses the aspects such as word choice and grammar in the language modality, tone and prosody in the audio modality, and facial expressions in the visual modality.
The \textit{disentangle and fuse} methods explicitly disentangle the representation of each modality into a modality-agnostic representation and modality-specific one, which are then aggregated into a multimodal joint representation. 
They aim towards bridging the differences of modality distribution within the modality-invariant subspace and \textit{explicitly} learning shared information, thus achieving promising performance and possessing good interpretability.

\begin{figure}[t]
	\centering
	\includegraphics[width=0.8\linewidth]{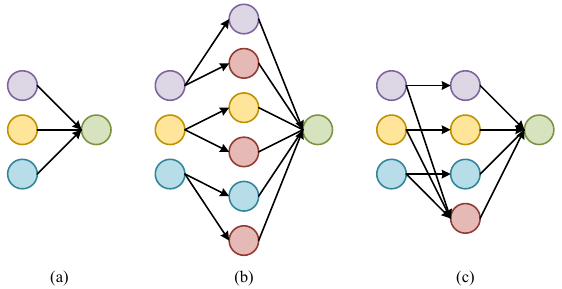}
	\caption{The illustration of the existed and our multimodal fusion methods. The subfigure (a), (b), and (c) denote the \textit{directly fuse}, previous \textit{disentangle and fuse} methods and our method, respectively. The purple, orange, and blue circles in the middle of subfigure (b) and (c) denote the modality-specific representations learned from visual, language, and audio modalities, respectively. The red circles denote the modality-agnostic representation. The green circles denote the multimodal joint representation.}
	\label{fig:PGM}
\end{figure}

However, existing representations disentanglement-based methods suffer from the following issues. 
On the one hand, in terms of model architecture, these methods first independently learn the modality-agnostic representations for each modality, and then aggregate these modality-agnostic representations together with their respective modality-specific representations to form a multimodal joint representation. 
On the other hand, in terms of information constraints, they enforce orthogonality between the modality-invariant and modality-specific subspace to reduce the linear correlation between corresponding representations, but overlook their non-linear correlations, thus failing to ensure the effective disentanglement of the representations for each modality.
Both of these issues lead to information redundancy in the multimodal joint representation, thus affecting the generalization capability of the model.
In this paper, we explore a superior paradigm to tackle the above issues in model architecture design and information constraint levels, respectively.

To address the first issue, we propose a novel representations disentanglement framework, as illustrated in Figure \ref{fig:PGM}.
Instead of separately learning modality-agnostic representations for each modality, we only learn a unified modality-agnostic representation. 
Then, we employ mutual information to measure the non-linear correlations between the latent representations.
By minimizing the mutual information between the modality-agnostic and modality-specific representations, as well as among the modality-specific representations themselves, we ensure information independence between corresponding representations, achieving the superior disentanglement of the representations for each modality and avoiding information redundancy within multimodal joint representation. 
We refer to our approach as Mutual Information-based Representations Disentanglement (MIRD). 
Last, we introduce unlabeled data to accurately estimate the mutual information.
Meanwhile, the use of unlabeled data is also capable of helping to characterize the underlying structure of multimodal data and further mitigates overfitting.

To validate the effectiveness of the proposed MIRD, extensive experiments were conducted on several benchmark datasets for multimodal sentiment analysis. 
The experimental results demonstrate that MIRD not only outperforms the baseline models but also achieves the state-of-the-art performance. 
In addition, the visualization of the disentangled representations further confirms that the proposed method effectively disentangles the representation for each modality.

Our contributions are as follows:
\begin{itemize}
	\item We propose a novel mutual information-based representations disentanglement method with the different framework from previous methods. It enables the superior disentanglement of modality-specific and modality-agnostic representations, avoiding information redundancy within multimodal joint representation.
	\item We introduce unlabeled data to accurately estimate the mutual information, which is also helpful to characterize the underlying structure of multimodal data to further improve the performance of the model.
	\item Comprehensive experiments on several benchmark datasets show that our model outperforms state-of-the-art methods by a large margin.
\end{itemize}

The rest of this paper is organized as follows. 
We review related works in Section \ref{sec:related_works}. 
In Section \ref{sec:method}, we present our mutual information-based representations disentanglement method. 
Experiments and analysis are provided in Section \ref{sec:experiment-settings} and \ref{sec:results}.
Finally, the conclusions are discussed in Section \ref{sec:conclusion}.

\section{Related Works}
\label{sec:related_works}

\subsection{Directly fuse}
\label{ssec:direct-fuse}
\textbf{Tensor-based methods} Although tensor-based methods were initially applied to aligned multimodal language sequences, they can be easily extended to the case of unaligned data.
TFN \cite{TFN} utilizes an outer product-based calculation of representations encoded from each modality to capture uni-modal, bi-modal, and tri-modal interactions.
LMF \cite{LMF} builds upon TFN by incorporating a low-rank multimodal tensor fusion technique, enhancing the efficiency of the model.

\textbf{Graph-based methods}  Multimodal Graph \cite{GPFN} is structured hierarchically for two stages: intra- and inter-modal dynamics learning. 
In the first stage, a graph convolutional network is used for each modality to learn intra-modal dynamics. 
In the second stage, a graph pooling fusion network is used to automatically learn associations between nodes from different modalities and form a multimodal joint representation.
GraphCAGE \cite{GraphCAGE} leverages the dynamic routing of capsule network and self-attention to create nodes and edges for each modality, which effectively address long-range dependency issues.
MTGAT \cite{MTGAT} presents a procedure for transforming unaligned multimodal sequence data into a single graph comprising heterogeneous nodes and edges. 
This graph effectively captures the rich interactions among diverse modalities through time. 
Subsequently, a novel multimodal temporal graph attention is introduced to incorporate a dynamic pruning and read-out technique, enabling the refined joint representation of this multimodal temporal graph.

\textbf{Cross-modal attention-based methods} MulT \cite{MulT} utilizes a cross-modal attention mechanism, which provides a latent cross-modal adaptation that fuses multimodal information by directly attending to low-level features in other modalities, regardless of the need for alignment.
PMR \cite{PMR} utilizes a message hub to facilitate information exchange among unaligned multimodal sequences. 
The message hub sends common messages to each modality and reinforces their features through cross-modal attention. 
In turn, it gathers the reinforced features from each modality to create a reinforced common message. 
Through iterative cycles, the common message and modality features progressively complement each other.
LMR-CBT \cite{LMR-CBT} introduces a novel cross-modal block transformer, enabling complementary learning across different modalities. 
It comprises local temporal learning, cross-modal feature fusion, and global self-attention representations.

\subsection{Disentangle and fuse}
\label{disentangle-fuse}
MISA \cite{MISA} projects each modality into two subspaces: one is modality-invariant subspace for capturing commonalities and reducing modality gaps, and another is modality-specific subspace for preserving unique characteristics. 
These representations are fused into a multimodal joint representation for sentiment prediction. 
FDMER \cite{FDMER} handles modality heterogeneity by learning two distinct representations for each modality on \textit{aligned} multimodal sequences.
The first representation is the common representation for projecting all modalities into a modality-invariant shared subspace with aligned distributions.
The representation can capture commonality among modalities regarding suggested emotions and reduces the modality gap.
The second representation is the private representation for providing a modality-specific subspace for each modality. 
In this subspace, FDMER learns the unique characteristics of different modalities and eliminates redundant information.
MFSA \cite{MFSA} employs a predictive self-attention module to capture reliable contextual dependencies and enhance unique features within modality-specific spaces. 
In addition, a hierarchical cross-modal attention module is introduced to explore correlations between cross-modal elements across the modality-agnostic space. 
To ensure the acquisition of distinct representations, a double-discriminator strategy is employed in an adversarial manner. 
Finally, modality-specific and modality-agnostic representations are integrated into a multimodal joint representation for downstream tasks.

\section{Methodology}
\label{sec:method}
\begin{figure*}[tp]
	\centering
	\includegraphics[width=\textwidth]{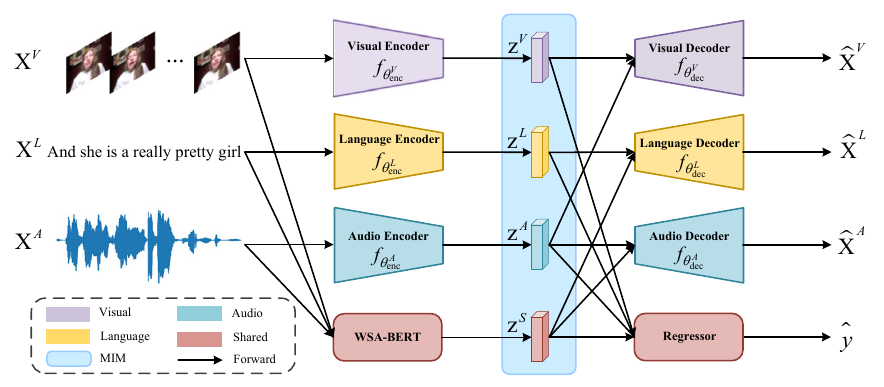}
	\caption{The framework of Mutual Information-based Representations Disentanglement (MIRD). The arrows indicate the forward calculation process. The blue shaded box indicate the Mutual Information Minimization (MIM) module.}
	\label{fig:framework}
\end{figure*}

In previous representations disentanglement-based frameworks, the feature sequence of each modality is separately processed through a modality-agnostic (a.k.a., shared/common) encoder to obtain the respective modality-agnostic representation. 
In contrast, we argue that learning modality-agnostic representations requires the cross-modal interactions, and obtaining modality-agnostic representations for each modality through a single modality-agnostic encoder is suboptimal and results in information redundancy in the aggregated multimodal joint representation.
In this section, we describe our proposed Mutual Information-based Representations Disentanglement (MIRD) framework, as shown in Figure \ref{fig:framework}.
The framework includes the encoders and decoders for each modality, the multimodal encoder, the Mutual Information Minimization (MIM) module, and the regressor.
In the subsequent subsections, we will describe these modules in detail.
\subsection{Formulation}
Our task is defined as follows: Given a set of three modalities $M=\{L ({\rm Language}),A ({\rm Audio}),V ({\rm Visual})\}$, an opinion video, i.e., multimodal language sequence, can be represented as $\{\mathbf{X}^{V},\mathbf{X}^{L},\mathbf{X}^{A}\}$ where $\mathbf{X}^{m}=[\mathbf{x}_{1}^{m},\mathbf{x}_{2}^{m},\\$$...,\mathbf{x}_{T_m}^{m}]\in \mathbb{R}^{T_m\times d_m}$, $m \in M-\{L\}$, $\mathbf{x}_i^{m}\in \mathbb{R}^{d_m}$ denotes the extracted sentiment features corresponding to modality $m$ at the $i$-th moment, $d_m$ is the dimension of the feature, and $T_m$ is the length of sequence of modality $m$. 
$\mathbf{X}^L=\left[w_1,w_2,...,w_{T_L}\right]$ is the original sequence of words.
Our goal is to predict the sentiment intensity $y \in \mathbb{R}$ of the whole video.

\subsection{Modality-specific representations learning}
To extract modality-specific (a.k.a., private) information, we design the unimodal encoders to map the feature sequences of each modality to their respective modality-specific subspaces, obtaining mutually exclusive modality-specific representations,

\begin{equation}
\label{equ:unimodal-encoder}
\mathbf{z}^m=f_{\theta_{\rm enc}^{m}}\left(\mathbf{X}^m\right)
\end{equation}
where $m \in M$, $f_{\theta_{\rm enc}^{m}}$ represents the encoder for modality $m$,
$\theta_{\rm enc}^{m}$ denotes the parameter of the encoder,
$\mathbf{z}^m \in \mathbb{R}^{d}$ is the acquired modality-specific representation, and $d$ is the dimension of the representation.

For all modalities, the Long Short-Term Memory (LSTM) \cite{LSTM} and subsequent fully connected networks are selected as the modality-specific encoders $f_{\theta_{\rm enc}^{L}},f_{\theta_{\rm enc}^{A}},f_{\theta_{\rm enc}^{V}}$ to model the intra-modal interactions within the sequences.

\subsection{Modality-agnostic representation learning}
The aim of the multimodal encoder is to obtain a modality-agnostic representation from the feature sequences of three modalities, and the learning of such a representation requires cross-modal interactions. 
Given the outstanding performance and popularity of text-centric cross-modal interaction models, we introduce our previous work Word-wise Sparse Attention BERT (WSA-BERT) \cite{WSA-BERT}, which takes into account long-range dependencies across modalities and sparse attention mechanisms between sequence elements.
By injecting the non-verbal (audio and visual) information into the word representations at the intermediate stage of the text backbone model, more refined word representations are obtained.

More specifically, we first feed $\mathbf{X}^L$ into a pretrained BERT \cite{BERT} model and extract the middle word representations $\mathbf{W}^{\rm mid}=\left[\mathbf{w}_1,\mathbf{w}_2,...,\mathbf{w}_{T_{\scriptscriptstyle L}}\right] \in \mathbb{R}^{T_L\times d_w}$, where $\mathbf{w}_k \in \mathbb{R}^{d_w} \left(k=1,2,..,T_L\right)$ is the intermediate word representation at the $k$-th moment, $d_w$ is the dimension of the representation, and $T_L$ represents the length of the sequence.
Then, the non-verbal feature sequences $\mathbf{X}^A$ and $\mathbf{X}^V$, along with the word representation sequence $\mathbf{W}^{\rm mid}$, are jointly mapped to a common semantic space, where the word $\mathbf{w}_k$ is used as a semantic anchor to search for the non-verbal information most relevant to $\mathbf{w}_k$ from holistic non-verbal sequences.
The semantic similarity between each word representation and the non-verbal features is calculated and then followed by a Sparsemax \cite{Sparsemax} function to highlight the important information in non-verbal sequences,
\begin{equation}
\label{equ:sparsemax}
\widetilde{\mathbf{X}}^m={\rm Sparsemax}\left(f_{\theta_{\rm sim}^{m}}\left(\mathbf{W}^{\rm mid},\mathbf{X}^m\right)\right)\cdot\mathbf{X}^m
\end{equation}
where $m \in M-\{L\}$, $f_{\theta_{\rm sim}^{m}}$ is the similarity function (fully connected network and inner product similarity) with the parameter $\theta_{\rm sim}^{m}$ for modality $m$,
${\rm Sparsemax}(\cdot)$ is the Sparsemax transformation which can map attention scores to a probability simplex,
$\widetilde{\mathbf{X}}^A \in \mathbb{R}^{T_L\times d_A}$ and $\widetilde{\mathbf{X}}^V \in \mathbb{R}^{T_L\times d_V}$ denote the combined audio and visual feature sequences with the same length as the language sequence $\mathbf{W}^{\rm mid}$, respectively.

Furthermore, a Multimodal Adaptive Gating (MAG) \cite{MAG-BERT} mechanism is leveraged to determine the amount of information injected from non-verbal modalities.
Specifically, two adaptive gates are constructed to control audio and visual information, respectively,
\begin{equation}
\label{equ:adaptive-gate}
\mathbf{G}^m=\sigma\left(f_{\theta_{\rm ada}^{m}}\left(\left[\mathbf{W}^{\rm mid};\widetilde{\mathbf{X}}^{m}\right]\right)\right)
\end{equation}
where $m \in M-\{L\}$, $\left[\mathbf{W}^{\rm mid};\widetilde{\mathbf{X}}^{m}\right]$ represents the concatenation of $\mathbf{W}^{\rm mid}$ and $\widetilde{\mathbf{X}}^{m}$, $\sigma$ represents a sigmoid activation function, $f_{\theta_{\rm ada}^{m}}$ is the single-layer fully connected network with the parameter $\theta_{\rm ada}^{m}$, and $\mathbf{G}^m \in \mathbb{R}^{T_L\times d_w}$ is the acquired adaptive gate.

Then, the sequence of displacement vectors is acquired by fusing together audio and visual features multiplied by their respective gate vectors,
\begin{equation}
\label{equ:displacement-vector}
\mathbf{H}=\mathbf{G}^V\odot f_{\theta_{\rm dis}^{V}}\left(\widetilde{\mathbf{X}}^{V}\right)+\mathbf{G}^A\odot f_{\theta_{\rm dis}^{A}}\left(\widetilde{\mathbf{X}}^{A}\right)
\end{equation}
where $\odot$ represents the Hadamard product (element-wise product), $f_{\theta_{\rm dis}^{V}}$ and $f_{\theta_{\rm dis}^{A}}$ are the single-layer fully connected networks with the parameters $\theta_{\rm dis}^{V}$ and $\theta_{\rm dis}^{A}$, respectively, $\mathbf{H} \in \mathbb{R}^{T_L\times d_w}$ is the acquired sequence of displacement vectors.
Subsequently, the weighted summation is utilized between BERT word representations $\mathbf{W}^{\rm mid}$ and its non-verbal displacement vectors $\mathbf{H}$ to create a more refined word representations sequence containing non-verbal information,
\begin{equation}
\label{equ:refined-repr}
\widetilde{\mathbf{W}}^{\rm mid}=\mathbf{W}^{\rm mid}+{\rm Diag}\left(\overrightarrow{\lambda}\right)\cdot\mathbf{H}
\end{equation}
\begin{equation}
\label{equ:scale-factor}
\overrightarrow{\lambda}=\left[\mathop{\min}\left(\frac{{\Vert \mathbf{w}_{1}\Vert}_2}{{\Vert \mathbf{h}_{1}\Vert}_2}\epsilon,1\right),...,\mathop{\min}\left(\frac{{\Vert \mathbf{w}_{T_{\scriptscriptstyle L}}\Vert}_2}{{\Vert \mathbf{h}_{T_{\scriptscriptstyle L}}\Vert}_2}\epsilon,1\right)\right]
\end{equation}
where $\epsilon$ is a hyper-parameter, $\lambda_k$ is the scaling factor which designed to remain the effect of non-verbal displacement vectors $\mathbf{h}_k$ within a desirable range, ${\rm Diag}(\cdot)$ is a function that transforms vectors into diagonal matrices, $\widetilde{\mathbf{W}}^{\rm mid} \in \mathbb{R}^{T_L\times d_w}$ is the modified word representation incorporating global non-verbal information.
We input $\widetilde{\mathbf{W}}^{\rm mid}$ into the subsequent BERT model to obtain the final word representations $\mathbf{W}^{\rm last} \in \mathbb{R}^{T_L\times d_w}$.
Finally, the word representations $\mathbf{W}^{\rm last}$ are pooled and mapped to the modality-agnostic subspace,
\begin{equation}
\label{equ:multimodal-encoder}
\mathbf{z}^S=f_{\theta_{\rm FC}}\left(\rm Pooling\left(\mathbf{W}^{\rm last}\right)\right)
\end{equation}
where $\rm Pooling\left(\cdot\right)$ is the average pooling function, $f_{\theta_{\rm FC}}$ denotes the fully connected network with the parameter $\theta_{\rm FC}$, $\mathbf{z}^{S} \in \mathbb{R}^{d}$ is the acquired modality-agnostic representation, and $d$ is the dimension of the representation. 
For simplicity, we define the parameter of the multimodal encoder as $\theta_{\rm enc}^{S}=\{\theta_{\rm sim}^{V},\theta_{\rm sim}^{A},\theta_{\rm ada}^{V},\theta_{\rm ada}^{A},\theta_{\rm dis}^{V},\theta_{\rm dis}^{A},\theta_{\rm FC}\}$.

\subsection{Multimodal reconstruction}
\label{ssec:decoder}
To ensure that the disentangled representations do not lose information and capture the underlying structure of multimodal data \cite{DAE}, the disentangled modality-specific representation $\mathbf{z}^m$ and modality-agnostic representation $\mathbf{z}^S$ are used to reconstruct the original representations $\mathbf{X}^{m}$,
\begin{equation}
\label{equ:decoder}
\widehat{\mathbf{X}}^{m}=f_{\theta_{\rm dec}^{m}}\left(\mathbf{z}^{m},\mathbf{z}^{S}\right)
\end{equation}
where $m \in M$, $f_{\theta_{\rm dec}^{m}}$ is the LSTM \cite{LSTM} decoder for modality $m$ with the parameter $\theta_{\rm dec}^{m}$, $\widehat{\mathbf{X}}^{m} \in \mathbb{R}^{T_m\times d_m}$ is the reconstructed sequence of representations.

\subsection{Mutual information minimization}
\label{ssec:MIM}
The representations disentanglement aims to decompose the original representation $\mathbf{X}^{m}$ into $\mathbf{z}^{m}$ and $\mathbf{z}^{S}$ $(m \in M)$, such that $\mathbf{z}^{m}$ and $\mathbf{z}^{S}$ contain completely different information.
This requires removing any correlation between $\mathbf{z}^{m}$ and $\mathbf{z}^{S}$, between $\mathbf{z}^{m_1}$ and $\mathbf{z}^{m_2}$ $(m_1,m_2 \in M, m_1 \ne m_2)$.
Previous representations disentanglement-based methods \cite{MISA,MFSA} have often utilized Orthogonal Constraints (OC), taking $\mathbf{z}^{m}$ and $\mathbf{z}^{S}$ as examples, as shown below:
\begin{equation}
\label{equ:oc}
\mathcal{L}_{\rm oc}^{(m,S)}=\left\Vert \mathbf{Z}^{m} \cdot {\mathbf{Z}^{S}}^\top \right\Vert_{F}^{2}
\end{equation}
where $m \in M$, $\mathbf{Z}^{m} \in \mathbb{R}^{N_{\rm oc}\times d}$ and $\mathbf{Z}^{S} \in \mathbb{R}^{N_{\rm oc}\times d}$ are sets of $\mathbf{z}^{m}$ and $\mathbf{z}^{S}$ from $N_{\rm oc}$ samples, respectively, $\left\Vert\right\Vert_{F}$ represents the Frobenius norm.

The application of orthogonal constraints ensures linear independence among variables. Nonetheless, nonlinear relationships may still persist among these variables, leading to an inadequate level of representations disentanglement.
To this end, we propose a constraint based on mutual information minimization, which effectively eliminates the dependency between variables.
Specifically, we utilize Contrastive Log-ratio Upper Bound (CLUB) \cite{CLUB} to estimate the mutual information upper bound.
CLUB bridges mutual information estimation with contrastive learning, where mutual information is estimated by the difference of conditional probabilities between the joint distribution (positive samples) and the marginal distribution (negative samples),
\begin{equation}
\begin{aligned}
\label{equ:mim}
\mathcal{I}\left(\mathbf{z}^{m},\mathbf{z}^{S}\right)=  \mathbb{E}_{p(\mathbf{z}^{m},\mathbf{z}^{S})}\left[{\rm log}p(\mathbf{z}^{S}|\mathbf{z}^{m})\right]-
\mathbb{E}_{p(\mathbf{z}^{m})p(\mathbf{z}^{S})}\left[{\rm log}p(\mathbf{z}^{S}|\mathbf{z}^{m})\right]
\end{aligned}
\end{equation}
With sample pairs $\{(\mathbf{z}_{i}^{m},\mathbf{z}_{i}^{S})\}_{i=1}^{N_{\rm mi}}$, $\mathcal{I}\left(\mathbf{z}^{m},\mathbf{z}^{S}\right)$ has an unbiased estimation as:
\begin{equation}
\begin{aligned}
\label{equ:mim-sample}
\widehat{\mathcal{I}}\left(\mathbf{z}^{m},\mathbf{z}^{S}\right)= \frac{1}{N_{\rm mi}}\sum_{i=1}^{N_{\rm mi}}{\rm log}p(\mathbf{z}_{i}^{S}|\mathbf{z}_{i}^{m})-
\frac{1}{N_{\rm mi}^2}\sum_{i=1}^{N_{\rm mi}}\sum_{j=1}^{N_{\rm mi}}{\rm log}p(\mathbf{z}_{j}^{S}|\mathbf{z}_{i}^{m})
\end{aligned}
\end{equation}
Since the conditional distribution $p(\mathbf{z}^{S}|\mathbf{z}^{m})$ is unavailable, the above formula is difficult to calculate.
Therefore, a variational distribution $q_{\theta_{\rm var}^{(m,S)}}(\mathbf{z}^{S}|\mathbf{z}^{m})$ with parameter $\theta_{\rm var}^{(m,S)}$ is used to approximate $p(\mathbf{z}^{S}|\mathbf{z}^{m})$,
\begin{equation}
\begin{aligned}
\label{equ:mim-variational}
\widehat{\mathcal{I}}\left(\mathbf{z}^{m},\mathbf{z}^{S}\right)= \frac{1}{N_{\rm mi}}\sum_{i=1}^{N_{\rm mi}}{\rm log}q_{\theta_{\rm var}^{(m,S)}}(\mathbf{z}_{i}^{S}|\mathbf{z}_{i}^{m})-
\frac{1}{N_{\rm mi}^2}\sum_{i=1}^{N_{\rm mi}}\sum_{j=1}^{N_{\rm mi}}{\rm log}q_{\theta_{\rm var}^{(m,S)}}(\mathbf{z}_{j}^{S}|\mathbf{z}_{i}^{m})
\end{aligned}
\end{equation}
In theory, as long as the variational distribution $q_{\theta_{\rm var}^{(m,S)}}(\mathbf{z}^{S}|\mathbf{z}^{m})$ is good enough, (\ref{equ:mim-variational}) remains an upper bound on the mutual information \cite{CLUB}.
We employ the neural networks $f_{\theta_{\rm var}^{(m,S)}}$ parameterized by $\theta_{\rm var}^{(m,S)}$ to characterize the variational distribution $q_{\theta_{\rm var}^{(m,S)}}(\mathbf{z}^{S}|\mathbf{z}^{m})$, and by enhancing the capacity of the neural network and updating its parameters, we can obtain a more accurate approximation of $p(\mathbf{z}^{S}|\mathbf{z}^{m})$.
Thus, the neural networks are referred to as the mutual information estimators.

To minimize the mutual information between $\mathbf{z}^{m}$ and $\mathbf{z}^{S}$, at each training iteration, we first update the variational approximation $q_{\theta_{\rm var}^{(m,S)}}(\mathbf{z}^{S}|\mathbf{z}^{m})$ by minimizing the negative log-likelihood,
\begin{equation}
\label{equ:lld}
\mathcal{L}_{\rm lld}^{(m,S)}=-\frac{1}{N_{\rm mi}}\sum_{i=1}^{N_{\rm mi}}{\rm log}(q_{\theta_{\rm var}^{(m,S)}}(\mathbf{z}_{i}^{S}|\mathbf{z}_{i}^{m}))
\end{equation}
After $q_{\theta_{\rm var}^{(m,S)}}(\mathbf{z}^{S}|\mathbf{z}^{m})$ is updated, we calculate the estimator as described in (\ref{equ:mim-variational}).
Finally, $q_{\theta_{\rm var}^{(m,S)}}(\mathbf{z}^{S}|\mathbf{z}^{m})$ and $\left(f_{\theta_{\rm enc}^{m}},f_{\theta_{\rm enc}^{S}}\right)$ are updated alternately during the training. 

\begin{figure}[h]
	\centering
	\includegraphics[width=0.6\linewidth]{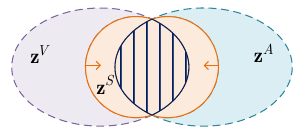}
	\caption{The diagram of the information constraints between the modality-agnostic representations and the original inputs. The purple and blue regions represent the information contained in the private representations $\mathbf{z}^{V}$ and $\mathbf{z}^{A}$ corresponding to visual and audio modalities, respectively. The orange region represents the information contained in the learned representation $\mathbf{z}^{S}$. The region with vertical stripes represents the \textit{genuine} shared information between visual and audio modalities. With the mutual information minimization constraints between $\mathbf{z}^{S}$ and $\mathbf{X}^{V}$, as well as $\mathbf{z}^{S}$ and $\mathbf{X}^{A}$, the area of the entire orange region gradually shrinks towards the area of the region with vertical stripes, indicating that $\mathbf{z}^{S}$ only contains shared information.}
	\label{fig:info}
\end{figure}

In addition to enforcing the independence of information between $\mathbf{z}^{V}$, $\mathbf{z}^{L}$, $\mathbf{z}^{A}$ and $\mathbf{z}^{S}$, it is also necessary to impose constraints on the independence of information between $\mathbf{z}^{S}$ and the original inputs $\mathbf{X}^{V}$, $\mathbf{X}^{L}$, $\mathbf{X}^{A}$ to truly learn the shared information between modalities. 
As shown in Figure \ref{fig:info}, taking visual and audio modalities as an example, without the constraint of information independence between $\mathbf{z}^{S}$ and the original inputs, $\mathbf{z}^{S}$ may contain excessive information beyond the genuine shared information.
Thus we further use variational distribution $q_{\theta_{\rm var}^{(Xm,S)}}(\mathbf{z}^{S}|\mathbf{X}^{m})$ to minimize the mutual information between $\mathbf{z}^{S}$ and $\mathbf{X}^{m}$,
\begin{equation}
\begin{aligned}
\label{equ:mim-variational2}
\widehat{\mathcal{I}}\left(\mathbf{X}^{m},\mathbf{z}^{S}\right)= &\frac{1}{N_{\rm mi}}\sum_{i=1}^{N_{\rm mi}}{\rm log}q_{\theta_{\rm var}^{(Xm,S)}}(\mathbf{z}_{i}^{S}|\mathbf{X}_{i}^{m})-\\&\frac{1}{N_{\rm mi}^2}\sum_{i=1}^{N_{\rm mi}}\sum_{j=1}^{N_{\rm mi}}{\rm log}q_{\theta_{\rm var}^{(Xm,S)}}(\mathbf{z}_{j}^{S}|\mathbf{X}_{i}^{m})
\end{aligned}
\end{equation}
The corresponding negative log-likelihood is as follows,
\begin{equation}
\label{equ:lld2}
\mathcal{L}_{\rm lld}^{(Xm,S)}=-\frac{1}{N_{\rm mi}}\sum_{i=1}^{N_{\rm mi}}{\rm log}(q_{\theta_{\rm var}^{(Xm,S)}}(\mathbf{z}_{i}^{S}|\mathbf{X}_{i}^{m}))
\end{equation}
For simplicity, we define the parameter of the variational distributions as $\theta_{\rm var}=\left\{\theta_{\rm var}^{(V,L)},\theta_{\rm var}^{(V,A)},\theta_{\rm var}^{(L,A)}, \theta_{\rm var}^{(V,S)}, \theta_{\rm var}^{(L,S)},\theta_{\rm var}^{(A,S)},\theta_{\rm var}^{(XV,S)},\theta_{\rm var}^{(XL,S)},\theta_{\rm var}^{(XA,S)}\right\}$.

\subsection{Sentiment prediction}
\label{ssec:regressor}
After obtaining the disentangled representations $(\mathbf{z}^V,\mathbf{z}^L,$
$\mathbf{z}^A,\mathbf{z}^S)$, a regressor $f_{\theta_{\rm reg}}$ is designed to predict the sentiment intensity of the video,
\begin{equation}
\label{equ:concatenation}
\mathbf{p}=\mathbf{z}^{V}\oplus\mathbf{z}^{L}\oplus\mathbf{z}^{A}\oplus\mathbf{z}^{S}
\end{equation}
\begin{equation}
\label{equ:regressor}
\widehat{y}=f_{\theta_{\rm reg}}\left(\mathbf{p}\right)
\end{equation}
where $\mathbf{p}\in \mathbb{R}^{4d}$ is the concatenation of $\mathbf{z}^{V},\mathbf{z}^{L},\mathbf{z}^{A}$ and $\mathbf{z}^{S}$, $f_{\theta_{\rm reg}}$ is a two-layer fully connected network with the parameter $\theta_{\rm reg}$, $\widehat{y}$ is the predicted sentiment intensity.
For simplicity, we define $\theta=\left\{\theta_{\rm enc}^{V},\theta_{\rm enc}^{L},\theta_{\rm enc}^{A},\theta_{\rm enc}^{S},\theta_{\rm dec}^{V},\theta_{\rm dec}^{L},\theta_{\rm dec}^{A},\theta_{\rm reg}\right\}$.

\subsection{Optimization objective}
\label{ssec:oo}
The task-specifc loss estimates the quality of prediction during training.
We use the Mean Squared Error (MSE) loss for the regression task,
\begin{equation}
\label{equ:reg-loss}
\mathcal{L}_{\rm reg}=log\left(\frac{1}{N_{\rm bs}}\sum_{i=1}^{N_{\rm bs}}\left(\widehat{y}_i-y_i\right)^2\right)
\end{equation}
where $N_{\rm bs}$ is the number of the samples in each batch, taking the logarithm of the MSE loss is done to avoid cumbersome weight hyper-parameter tuning and balance the gradients of each loss during the optimization process \cite{MTL}.
For reconstruction of original representations, the MSE loss is utilized for audio and visual modalities, and the Cross Entropy (CE) loss is utilized for language modality ( sentences are composed of discrete words in the lexicon),
\begin{equation}
\resizebox{0.88\hsize}{!}{$
	\begin{aligned}
	\label{equ:recon-loss}
	\mathcal{L}_{\rm recon}=\;&log\left(\mathcal{L}_{\rm recon}^{V}\right)+log\left(\mathcal{L}_{\rm recon}^{L}\right)+log\left(\mathcal{L}_{\rm recon}^{A}\right)\\
	=\;&log\left({\rm MSE}\left(\widehat{\mathbf{X}}^{V},\mathbf{X}^{V}\right)\right)+log\left( {\rm CE}\left(\widehat{\mathbf{X}}^{L},\mathbf{X}^{L}\right)\right)+\\
	&log\left( {\rm MSE}\left(\widehat{\mathbf{X}}^{A},\mathbf{X}^{A}\right)\right)
	\end{aligned}$}
\end{equation}
Similarly, taking the logarithm of the reconstruction loss for each modality is done to automatically balance the gradients of each loss during the optimization process and avoid manual adjustment of the loss weights.
Furthermore, we define $\widehat{\mathcal{I}}\left(\mathbf{z}^{m},\mathbf{z}^{S}\right)$ as $\mathcal{L}_{\rm mim}^{(m,S)}$, $\widehat{\mathcal{I}}\left(\mathbf{z}^{m_1},\mathbf{z}^{m_2}\right)$ as $\mathcal{L}_{\rm mim}^{(m_1,m_2)}$, $\widehat{\mathcal{I}}\left(\mathbf{X}^{m},\mathbf{z}^{S}\right)$ as $\mathcal{L}_{\rm mim}^{(Xm,S)}$ $\left(m,m_1,m_2\in M\right)$, thus the Mutual Information Minimization (MIM) loss is as follows:
\begin{equation}
\label{equ:mim-loss}
\mathcal{L}_{\rm mim}=\sum_{\overline{m}_1,\overline{m}_2\in \overline{M}}\mathcal{L}_{\rm mim}^{(\overline{m}_1,\overline{m}_2)}+\sum_{m\in M}\mathcal{L}_{\rm mim}^{(Xm,S)}
\end{equation}
where $\overline{M}=M+\{S\}=\{L,A,V,S\}$, $\overline{m}_1\ne \overline{m}_2$. 
The overall learning of the model is performed by minimizing:
\begin{equation}
\label{equ:loss}
\mathcal{L}=\mathcal{L}_{\rm reg}+\mathcal{L}_{\rm recon}+\alpha\mathcal{L}_{\rm mim}
\end{equation}
where $\alpha$ is the weight coefficient.
Since the estimated mutual information values may become negative, we do not take their logarithm. 
Instead, we simply set the weights of the mutual information minimization losses to be the same.
Meanwhile, as mentioned in Section \ref{ssec:MIM}, before optimizing $\mathcal{L}$, we need to minimize the negative log-likelihood $\mathcal{L}_{\rm lld}$ for accurate mutual information estimation,
\begin{equation}
\label{equ:lld-loss}
\mathcal{L}_{\rm lld}=\sum_{\overline{m}_1,\overline{m}_2\in \overline{M}}\mathcal{L}_{\rm lld}^{(\overline{m}_1,\overline{m}_2)}+\sum_{m\in M}\mathcal{L}_{\rm lld}^{(Xm,S)}
\end{equation}
where $\overline{M}=M+\{S\}=\{L,A,V,S\}$, $\overline{m}_1\ne \overline{m}_2$.
$\mathcal{L}$ and $\mathcal{L}_{\rm lld}$ are optimized alternately during the training.

However, the estimation of mutual information may not be accurate due to limited labeled data availability \cite{CLUB}, and introducing additional data can improve this.
Thus, given the abundant supply of unlabeled opinion videos accessible on the Internet, we leverage them to assist in computing mutual information.
On the other hand, these unlabeled data is also utilized to reconstruct original inputs, thereby aiding in characterizing the underlying structure of multimodal data, and further improving the performance of the model.
The overall optimization algorithm of our model is shown in Algorithm \ref{alg:optimization},

\begin{algorithm}[H]
	\caption{Optimization Process of MIRD}
	\label{alg:optimization}
	\begin{algorithmic}
		\STATE 
		\STATE \textbf{Input: }Labeled data $\mathbf{X}^{\rm labeled}=\left\{\mathbf{X}_{i}^V,\mathbf{X}_{i}^L,\mathbf{X}_{i}^A\right\}_{i=1}^{N_{\rm bs}},$ Labels $\mathbf{Y}=\{y_i\}_{i=1}^{N_{\rm bs}},$ Unlabeled data $\mathbf{X}^{\rm unlabeled}=\left\{\mathbf{X}_{i}^V,\mathbf{X}_{i}^L,\mathbf{X}_{i}^A\right\}_{i=N_{\rm bs}+1}^{N_{\rm bs}+N_u},$ Initial parameters $\theta,\theta_{\rm var},$ Learning rate $ \eta_1,\eta_2, $ Iterations $ T,T', $ weight coefficient $\alpha$
		\STATE \hspace{0.5cm}$\textbf{for } t \in [1,2,...,T] \textbf{ do}$
		\STATE \hspace{0.85cm} Calculate $\left\{\mathbf{z}_{i}^{V},\mathbf{z}_{i}^{L},\mathbf{z}_{i}^{A},\mathbf{z}_{i}^{S}\right\}_{i=1}^{N_{\rm bs}+N_u}$ by using labeled and unlabeled data $\mathbf{X}^{\rm labeled}\cup\mathbf{X}^{\rm unlabeled}$
		\STATE \hspace{1cm}$\textbf{for } t' \in [1,2,...,T'] \textbf{ do}$
		\STATE \hspace{1.4cm} Calculate $\mathcal{L}_{\rm lld}$ by equation (\ref{equ:lld-loss}) with $\left\{\mathbf{z}_{i}^{V},\mathbf{z}_{i}^{L},\mathbf{z}_{i}^{A},\mathbf{z}_{i}^{S}\right\}_{i=1}^{N_{\rm bs}+N_u}$
		\STATE \hspace{1.6cm}$\theta_{\rm var} \gets \theta_{\rm var}-\eta_1\frac{\partial \mathcal{L}_{\rm lld}}{\partial \theta_{\rm var}}$
		\STATE \hspace{1cm}$\textbf{done}$
		\STATE \hspace{0.88cm} Calculate $\mathcal{L}_{\rm reg}$ by equation (\ref{equ:reg-loss}) with $\left\{\mathbf{z}_{i}^{V},\mathbf{z}_{i}^{L},\mathbf{z}_{i}^{A},\mathbf{z}_{i}^{S}\right\}_{i=1}^{N_{\rm bs}}\cup\mathbf{Y}$
		\STATE \hspace{0.88cm} Calculate $\mathcal{L}_{\rm recon}$ by equation (\ref{equ:recon-loss}) with $\left\{\mathbf{z}_{i}^{V},\mathbf{z}_{i}^{L},\mathbf{z}_{i}^{A},\mathbf{z}_{i}^{S}\right\}_{i=1}^{N_{\rm bs}+N_u}$
		\STATE \hspace{0.88cm} Calculate $\mathcal{L}_{\rm mim}$ by equation (\ref{equ:mim-loss}) with $\left\{\mathbf{z}_{i}^{V},\mathbf{z}_{i}^{L},\mathbf{z}_{i}^{A},\mathbf{z}_{i}^{S}\right\}_{i=1}^{N_{\rm bs}+N_u}$
		\STATE \hspace{0.88cm} Calculate $\mathcal{L}=\mathcal{L}_{\rm reg}+\mathcal{L}_{\rm recon}+\alpha\mathcal{L}_{\rm mim}$
		\STATE \hspace{1.05cm}$\theta \gets \theta-\eta_2\frac{\partial \mathcal{L}}{\partial \theta}$
		\STATE \hspace{0.5cm}$\textbf{done}$
		\STATE \textbf{Output: }$\theta^{*},\theta_{\rm var}^{*}$
	\end{algorithmic}
\end{algorithm}

\section{Experiment Settings}
\label{sec:experiment-settings}
\subsection{Datasets}
\label{ssec:datasets}
We conduct the evaluation of our methods on two extensively used datasets, namely CMU-MOSI \cite{MOSI} and CMU-MOSEI \cite{MOSEI}, which encompass three modalities: visual, language, and audio, to convey the sentiment of the speakers.
In addition, a sentiment-unlabeled dataset EmoVoxCeleb \cite{EmoVoxCeleb} is introduced for accurately estimating mutual inforamtion and reconstructing original inputs.

\textbf{CMU-MOSI} dataset comprises 93 opinion videos extracted from YouTube movie reviews. 
Each video is composed of multiple opinion segments, with each segment annotated with sentiment scores ranging from -3 (strong negative) to +3 (strong positive). 
To align with previous studies \cite{MulT,Self-MM,MFSA}, we employ 1,284 segments for training, 229 segments for validation, and 686 segments for testing.

\textbf{CMU-MOSEI} dataset, currently the largest multimodal sentiment analysis dataset, contains 22,846 movie review video clips sourced from YouTube. 
Each video clip is assigned a sentiment score between -3 (strong negative) and +3 (strong positive). 
To align with previous studies \cite{MulT,Self-MM,MFSA}, we utilize 16,326 video clips for training, 1,861 video clips for validation, and 4,659 video clips for testing. 

\textbf{EmoVoxCeleb} comprises a vast collection of over 100,000 video clips featuring over 1,200 speakers, also sourced from open-source media YouTube. 
The dataset exhibits a balanced distribution across genders and encompasses speakers with diverse ethnicities, accents, professions, and ages. 
Through a rough screening process, non-English video clips were excluded, resulting in the selection of 132,708 English video clips, representing 1,105 speakers.
We select a total of 81,630 video clips, which is five times the size of the training set in CMU-MOSEI, as the unlabeled data.

\subsection{Sentiment features}
\label{ssec:sentiment-feature}
The sentiment-related features are extracted for non-verbal (visual and audio) modalities.

\textbf{Visual:} The MultiComp OpenFace2.0 toolkit \cite{Openface} is employed to extract a comprehensive set of visual features, encompassing 340-dimensional facial landmarks, 35-dimensional facial action units, 6-dimensional head pose and orientation, 40-dimensional rigid and non-rigid shape parameters, and 288-dimensional eye gaze \footnote{For more details, you can see \url{https://github.com/TadasBaltrusaitis/OpenFace/wiki/Output-Format}}.

\textbf{Audio:} The librosa library \cite{librosa} is utilized for extracting frame-level acoustic features which include 1-dimensional logarithmic fundamental frequency (log F0), 20-dimensional Mel-Frequency Cepstral Coefficients (MFCCs), and 12-dimensional Constant-Q chromatogram (CQT).
These features are related to emotions and tone of speech according to \cite{CH-SIMS}.

\subsection{Training details}
\label{ssec:training-details}
All models are developed using the PyTorch toolbox \cite{Pytorch} and trained on NVIDIA RTX 3090 GPUs. 
We employ the AdamW \cite{AdamW} optimizer with an initial learning rate of $1e-5$ for the encoders, decoders, and the regressor. 
For the MIM module, we utilize the Adam \cite{Adam} optimizer with an initial learning rate of $1e-3$. 
The batch size $N_{\rm bs}$ is set to 32, and the training process runs for 100 epochs. 
The mutual information estimator is a two-layer fully connected network with a hidden layer dimension of 32.
The number $N_{\rm mi}$ of sample pairs for mutual information estimator is equal to batch size $N_{\rm bs}$, and the number $T'$ of iterations is set to 5. 
The hyper-parameter $\epsilon$ in MAG is set to 1, and
the dimension $d$ of the latent representations is 64.
All our experiments are conducted using the exact same random seed. 
In addition, we utilize the designated validation sets of CMU-MOSI and CMU-MOSEI datasets to identify the best hyper-parameters for our models.

\begin{table}[h]
	\centering
	\caption{\centering Performance on CMU-MOSI and CMU-MOSEI Datasets.}
	\label{tab:comparison-both}
	\resizebox{\textwidth}{36mm}{
	\begin{tabular}{c|*4{c}|*4{c}}
		\toprule[1pt]
		\textbf{Datasets} &\multicolumn{4}{c|}{\textbf{CMU-MOSI}} &\multicolumn{4}{c}{\textbf{CMU-MOSEI}} \\
		\textbf{Metric} &\textbf{Acc} $\uparrow$ &\textbf{F1} $\uparrow$ &\textbf{MAE} $\downarrow$ &\textbf{Corr} $\uparrow$ &\textbf{Acc} $\uparrow$ &\textbf{F1} $\uparrow$ &\textbf{MAE} $\downarrow$ &\textbf{Corr} $\uparrow$ \\
		\hline
		\rule{0pt}{1.0em}TFN (B)$^\Diamond$ \cite{TFN} &80.80 &80.70 &0.901 &0.698 &82.50 &82.10 &0.593 &0.700 \\
		\rule{0pt}{1.0em}LMF (B)$^\Diamond$ \cite{LMF} &82.50 &82.40 &0.917 &0.695 &82.00 &82.10 &0.623 &0.677 \\
		\rule{0pt}{1.0em}MulT$^*$ \cite{MulT} &80.45 &80.47 &0.892 &0.667 &81.02 &80.98 &0.605 &0.670 \\
		\rule{0pt}{1.0em}$\rhd$MISA (B)$^*$ \cite{MISA} &81.20 &81.17 &0.825 &0.722 &82.70 &82.67 &0.589 &0.703 \\
		\rule{0pt}{1.0em}GPFN$^\S$ \cite{GPFN} &80.60 &80.50 &0.933 &0.684 &81.40 &81.70 &0.608 &0.675 \\
		\rule{0pt}{1.0em}GraphCAGE$^\S$ \cite{GraphCAGE} &82.10 &82.10 &0.933 &0.684 &81.70 &81.80 &0.609 &0.670 \\
		\rule{0pt}{1.0em}PMR$^*$ \cite{PMR} &81.33 &81.30 &0.875 &0.669 &82.12 &82.07 &0.614 &0.675 \\
		\rule{0pt}{1.0em}LMR-CBT$^*$ \cite{LMR-CBT} &80.42 &80.38 &0.901 &0.657 &80.75 &80.79 &0.634 &0.653 \\
		\rule{0pt}{1.0em}Self-MM$^*$ \cite{Self-MM} &85.21 &85.18 &0.773 &0.774 &84.07 &84.12 &0.556 &0.750 \\
		\rule{0pt}{1.0em}WSA-BERT$^*$ \cite{WSA-BERT} &83.99 &83.83 &0.766 &0.779 &84.36 &84.39 &0.556 &0.763 \\
		\rule{0pt}{1.0em}$\rhd$MFSA (B)$^*$ \cite{MFSA} &83.32 &83.30 &0.804 &0.746 &83.92 &83.87 &0.566 &0.743 \\
		\rule{0pt}{1.0em}MIRD$^{1}$ (ours) &85.82 &85.74 &0.723 &0.799 &85.62 &85.60 &0.556 &0.781 \\
		\rule{0pt}{1.0em}MIRD$^{2}$ (ours) &\textbf{86.23} &\textbf{86.16} &\textbf{0.721} &\textbf{0.804} &\textbf{86.34} &\textbf{86.29} &\textbf{0.553} &\textbf{0.823} \\
		\bottomrule[1pt]
	\end{tabular}}
\end{table}

\subsection{Evaluation metrics}
Consistent with previous studies \cite{MulT,Self-MM,MFSA}, we record our experimental results in two forms: classification and regression.
For classification, we report the weighted F1 score and binary accuracy.
For regression, we report Mean Absolute Error (MAE) and Pearson correlation (Corr).
Except for MAE, higher values indicate better performance for all metrics.

\section{Results and Analysis}
\label{sec:results}
In this section, we make a detailed analysis and discussion about our experimental results.
All codes are publicly available at here \footnote{\url{https://github.com/qianfan1996/MIRD}}.
\subsection{Comparison with state-of-the-art models}
\label{ssec:comparison-results}
The comparison results with state-of-the-art models on the CMU-MOSI and CMU-MOSEI datasets are shown in Table \ref{tab:comparison-both}.
$\uparrow$ means higher is better, and $\downarrow$ means lower is better. 
(B) means the language features are based on BERT \cite{BERT}. $^\Diamond$ means the results are from \cite{Self-MM}; $\S$ from original papers. $^*$ denotes the reimplementation with non-verbal sentiment features mentioned in Section \ref{ssec:sentiment-feature}. $\rhd$ denotes other representations disentanglement-based methods.
MIRD$^{1}$ and MIRD$^{2}$ indicate the non-use and use of unlabeled data, respectively.
\% is omitted after Acc and F1 metrics.
Best results are highlighted in bold.

From the results we can observe that, 1) our method MIRD (with or without unlabeled data) performs much better than all compared methods.
The Accuracy, F1 score, MAE, and Correlation metrics of MIRD with unlabeled data are 86.23\%, 86.16\%, 0.721, and 0.804 on CMU-MOSI dataset, 86.34\%, 86.29\%, 0.553, and 0.823 on CMU-MOSEI dataset.
The absolute improvements of the performance are 1.02\%, 0.98\%, 0.052, and 0.030 on CMU-MOSI dataset, 2.27\%, 2.17\%, 0.003, and 0.073 on CMU-MOSEI dataset compared to the
second-best method Self-MM \cite{Self-MM}.
2) MIRD outperforms previous representations disentanglement-based methods, MISA \cite{MISA} and MFSA \cite{MFSA}, on both datasets (we do not make comparisons with \cite{DISRFN} and \cite{FDMER} since both of works are conducted on \textit{aligned} multimodal language sequences, and official code implementations are not provided).
These observations indicate that MIRD is more effective in modeling unaligned multimodal language sequences.

\subsection{Efficiency analysis}
\label{ssec:efficiency-analysis}
To analyze the efficiency of MIRD, we list the number of parameters (\#Params) and computation (Computational Budget) of each model, as shown in Table \ref{tab:num-param}.
Same as Table \ref{tab:comparison-both}, (B) means the language features are based on BERT \cite{BERT}; $\rhd$ denotes other representations disentanglement-based methods.
The M stands for millions; The GMACs denotes Giga Multiply–Accumulate Operations.
From the table we can find that the model efficiency of MIRD is also remarkable under the premise of the best performance, especially compared with other representations disentanglement-based methods.
This indicates that our method has relatively high applicability.
\begin{table}[h]
	\centering
	\caption{\centering The efficiency of the models.}
	\label{tab:num-param}
	\begin{tabular}{c|cc}
		\toprule[1pt]
		\textbf{Method} &\textbf{\#Params} &\textbf{Computational Budget}\\
		\hline
		\rule{0pt}{1.0em}$\rhd$MISA (B) &123 M  &5.0 GMACs\\
		\rule{0pt}{1.0em}Self-MM &112 M  &4.6 GMACs\\
		\rule{0pt}{1.0em}WSA-BERT &115 M  &4.7 GMACs\\
		\rule{0pt}{1.0em}$\rhd$MFSA (B) &138 M  &5.6 GMACs\\
		\rule{0pt}{1.0em}MIRD &121 M  &4.9 GMACs\\
		\bottomrule[1pt]
	\end{tabular}
\end{table}

\subsection{Comparison with other representations disentanglement-based methods}
\label{ssec:comparison-RD}
To illustrate the differences between our method and previous representations disentanglement-based methods, we conduct a comparison in Table \ref{tab:comparison-structure}.
In the table, \ding{51} denote presence, while \ding{55} indicate absence.
The difference between the multimodal encoder and the common encoder is that the input of the multimodal encoder is multiple modalities together, while the input of the common encoder is one of each modality.
Our method involves learning latent representations through cross-modal interactions, followed by a straightforward concatenation, whereas other representations disentanglement-based methods directly learn latent representations and then use the Transformer \cite{Transformer} for fusion.

\begin{table}[h]
	\centering
	\caption{\centering Comparison with other representations disentanglement-based methods.}
	\label{tab:comparison-structure}
	\begin{tabular}{c|*3{c}}
		\toprule[1pt]
		\textbf{Module} &\textbf{MIRD} &\textbf{MISA} &\textbf{MFSA} \\
		\hline
		\rule{0pt}{1.0em}Unimodal Encoder &\ding{51} &\ding{51} &\ding{51} \\
		\rule{0pt}{1.0em}Multimodal Encoder &\ding{51} &\ding{55} &\ding{55} \\
		\rule{0pt}{1.0em}Common Encoder &\ding{55} &\ding{51} &\ding{51} \\
		\rule{0pt}{1.0em}Decoder &\ding{51} &\ding{51} &\ding{55} \\
		\rule{0pt}{1.0em}Modality Discriminator &\ding{55} &\ding{55} &\ding{51} \\
		\rule{0pt}{1.0em}Non-linear Constraint &\ding{51} &\ding{55} &\ding{55} \\
		\rule{0pt}{1.0em}Fusion Mode &Concat &Transformer &Transformer \\
		\rule{0pt}{1.0em}Interact before learning &\ding{51} &\ding{55} &\ding{55} \\
		\bottomrule[1pt]
	\end{tabular}
\end{table}

\subsection{Ablation study}
\label{ssec:ablation-study}
To further explore the contributions of different components, we conduct an ablation study on CMU-MOSI dataset and the results are shown in Table \ref{tab:ablation-mosi}.
From this table, we can observe:

1) Experiment 1 removes the labeled data and uses only labeled data.
The performance decreases significantly (accuracy drops from 86.23\% to 85.82\%, F1 score drops from 86.16\% to 85.74\%), highlighting the importance of utilizing more data to estimate the mutual information and helping to characterize the underlying structure of the multimodal data.
However, despite the performance degradation observed in our method without unlabeled data, comparing it with other methods in Table \ref{tab:comparison-both}, our method still significantly outperforms the alternatives. 
This shows the effectiveness of the proposed disentanglement architecture along with information constraints.

2) Experiment 2, which omits the mutual information minimization constraint between the shared representation $\mathbf{z}^S$ and the input $\mathbf{X}^m$, does not result in a degradation of model performance.
This is because after the representations disentanglement, there is an aggregation process on the disentangled representations, and whether or not the mutual information minimization constraint is applied does not affect the information integrity of the aggregated multimodal joint representation $\mathbf{z}$.

3) Experiment 3 removes the mutual information minimization constraints among the latent representations $\mathbf{z}^{V},\mathbf{z}^{L},\mathbf{z}^{A}$ and $\mathbf{z}^{S}$.
The performance further deteriorates (accuracy drops from 85.80\% to 85.21\%, F1 score drops from 85.77\% to 85.09\%), indicating that explicitly learning modality-specific and modality-agnostic representations is beneficial for sentiment prediction.

4) Experiment 4 indicates that if the original inputs are not reconstructed and only the WSA-BERT model \cite{WSA-BERT} is used for sentiment prediction, the performance of the model decreases to an accuracy of 83.99\% and an F1 score of 83.83\%, further demonstrating the effectiveness of the proposed method.

\begin{table}[h]
	\centering
	\caption{\centering Ablation study on CMU-MOSI dataset.}
	\label{tab:ablation-mosi}
	\begin{tabular}{l|*4{c}}
		\toprule[1pt]
		\textbf{Method} &\textbf{Acc} $\uparrow$ &\textbf{F1} $\uparrow$ &\textbf{MAE} $\downarrow$ &\textbf{Corr} $\uparrow$ \\
		\hline
		\rule{0pt}{1.0em}MIRD$^{2}$ &\textbf{86.23\%} &\textbf{86.16\%} &\textbf{0.721} &\textbf{0.804} \\
		1. \rule{0pt}{1.0em}MIRD$^{1}$ &85.82\% &85.74\% &0.723 &0.799 \\
		2. \rule{0pt}{1.0em}$- \mathcal{L}_{\rm mim}^{(Xm,S)}$ &85.80\% &85.77\% &0.722 &0.796 \\
		3. \rule{0pt}{1.0em}$- \mathcal{L}_{\rm mim}$ &85.21\% &85.09\% &0.750 &0.787 \\
		4. \rule{0pt}{1.0em}$- \mathcal{L}_{\rm recon}$ &83.99\% &83.83\% &0.766 &0.779 \\
		\bottomrule[1pt]
	\end{tabular}
\end{table}

\subsection{Comparison between MIM and OC}
\label{ssec:mim-oc}
\begin{figure*}[h]
	\centering
	\subfigure[NC]{
		\includegraphics[trim = 0mm 0mm 0mm 4mm, clip=true, width=0.3\linewidth]{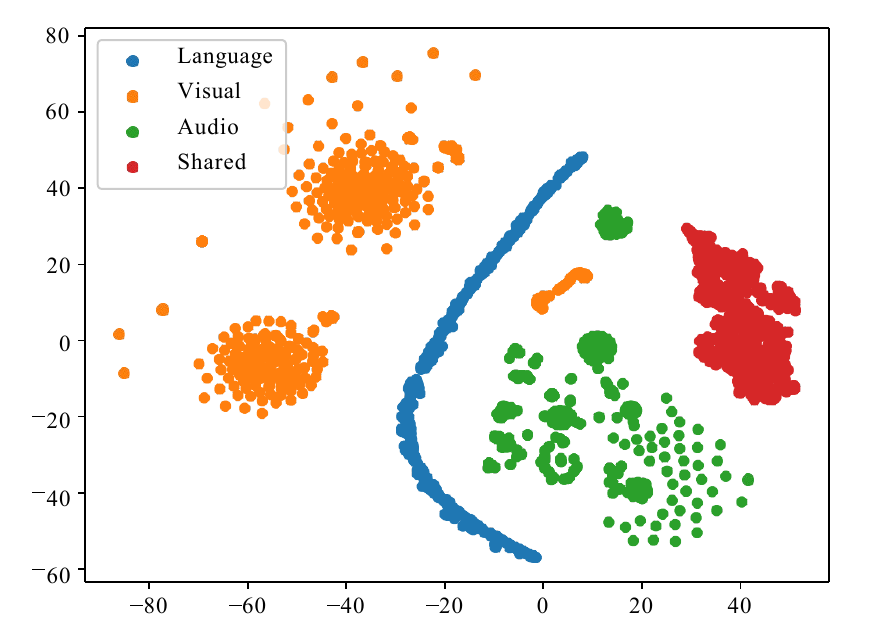}
	}
	\hfill
	\subfigure[OC]{
		\includegraphics[trim = 0mm 0mm 0mm 4mm, clip=true, width=0.3\linewidth]{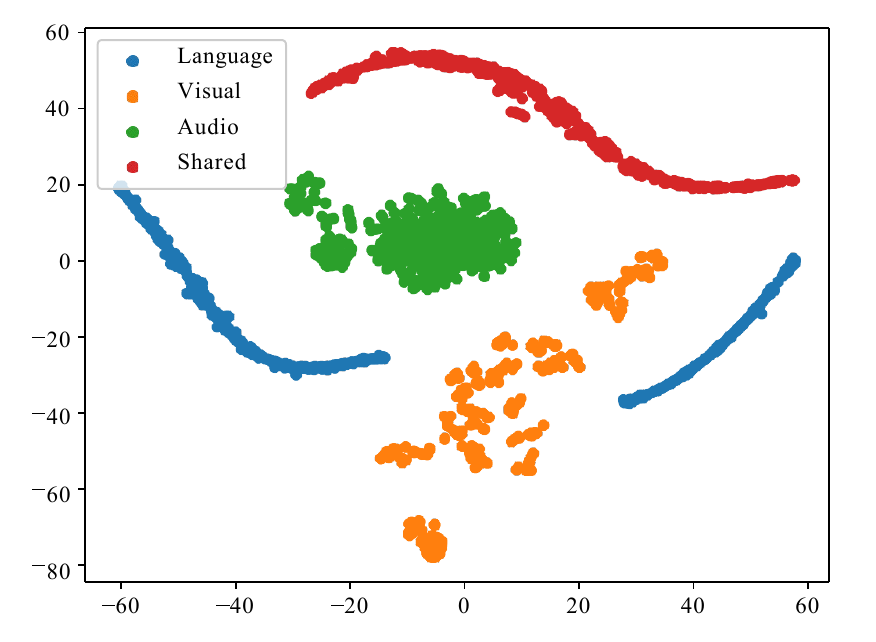}
	}
	\hfill
	\subfigure[MIM]{
		\includegraphics[trim = 0mm 0mm 0mm 4mm, clip=true, width=0.3\linewidth]{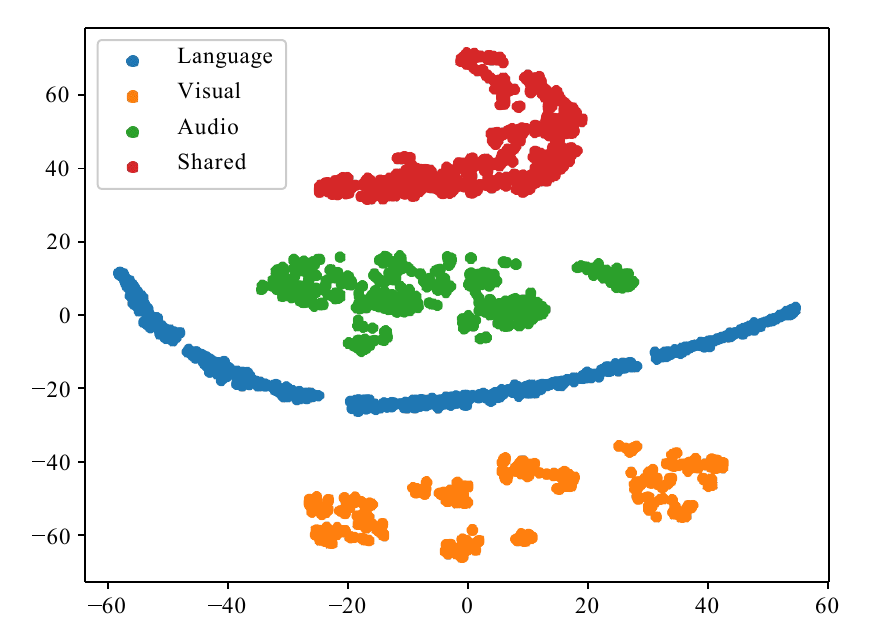}
	}
	\caption{Visualization of modality-agnostic and modality-specific representations on CMU-MOSI test set. The subfigure (a), (b), and (c) denote the No Constraints (NC), Orthogonal Constraints (OC), and Mutual Information Minimization constraint (MIM), respectively. In each subfigure, blue, orange, green, and red points represent language, visual, audio, and shared representations, respectively.}
	\label{fig:visualization}
\end{figure*}

The utilization of Orthogonal Constraint (OC) can enhance the linear independence between variables. 
However, it is possible that non-linear relationships still exist among them, leading to inadequate disentanglement of representations.
The representations disentanglement method based on Mutual Information Minimization (MIM) can effectively eliminate the dependencies between variables, thereby achieving a more comprehensive disentanglement of representations across different modalities. 
This reduces information redundancy in the multimodal joint representation and consequently enhances the performance of the model.
To validate this, we compare the performance of two constrained methods and visualize their disentangled representations, as shown in Table \ref{tab:comparison-both-datasets} and Figure \ref{fig:visualization}.

\begin{table}[h]
	\centering
	\caption{\centering Comparison experiments on both datasets.}
	\label{tab:comparison-both-datasets}
	\begin{tabular}{c|c|*4{c}}
		\toprule[1pt]
		\textbf{Dataset} &\textbf{Method} &\textbf{Acc} $\uparrow$ &\textbf{F1} $\uparrow$ &\textbf{MAE} $\downarrow$ &\textbf{Corr} $\uparrow$ \\
		\hline
		\multicolumn{1}{c|}{\multirow{2}[4]{*}[+1ex]{CMU-MOSI}} \rule{0pt}{1.0em}&OC &85.35\% &85.32\% &0.744 &0.790 \\
		\rule{0pt}{1.0em}&MIM &85.80\% &85.77\% &0.722 &0.796 \\
		\hline
		\hline
		\multicolumn{1}{c|}{\multirow{2}[4]{*}[+1ex]{CMU-MOSEI}} \rule{0pt}{1.0em}&OC &85.46\% &85.39\% &0.577 &0.794 \\
		\rule{0pt}{1.0em}&MIM &85.87\% &85.82\% &0.570 &0.798 \\
		\bottomrule[1pt]
	\end{tabular}
\end{table}

From Table \ref{tab:comparison-both-datasets}, we can observe that the  MIM-based method outperforms the OC-based method in almost all evaluation metrics on the CMU-MOSI and CMU-MOSEI datasets.
Specifically, on the CMU-MOSI dataset, the MIM-based method outperforms the OC-based method by 0.45\% in terms of both accuracy and F1 score; on the CMU-MOSEI dataset, the MIM-based method exhibits superior accuracy and F1 score compared to the OC-based method, with differences of 0.41\% and 0.43\%, respectively.
From Figure \ref{fig:visualization} (a), we can observe that in the method without relationship constraints, the disentangled visual representations (orange points) are separated from the language representations (blue points), and the major portion of the visual representations spontaneously form two distinct clusters.
Furthermore, the disentangled audio representations (green points) also exhibit a rough formation of two clusters.
From Figure \ref{fig:visualization} (b), it can be seen that in the OC-based method, the disentangled language representations (blue points) are separated from the visual representations (orange points).
However, the MIM-based method tends to roughly cluster language, visual, audio, and shared representations separately, demonstrating greater distinctiveness (as depicted in Figure \ref{fig:visualization} (c)).
It suggests that imposing relationship constraints on the latent representations yields more discriminative representations, with the MIM-based method showing superior distinctiveness compared to the OC-based method.
Interestingly, in Figure \ref{fig:visualization} (a), we observe that despite the absence of any relationship constraint, the intermediate representations obtained by the model still exhibit some level of separability.
This could be attributed to the fact that the individual modality encoders as well as the multimodal encoder have learned certain discriminative information.

\begin{figure}[t]
	\centering
	\includegraphics[trim = 0mm 0mm 0mm 2mm, clip=true, width=0.8\linewidth]{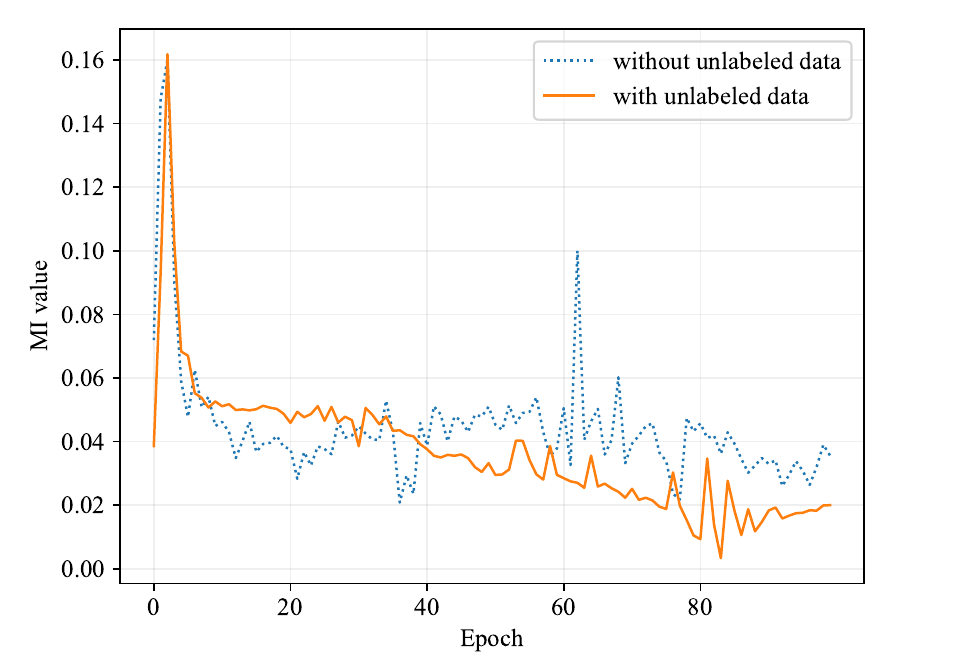}
	\caption{The line chart of mutual information estimation on CMU-MOSI dataset. The horizontal axis represents epochs, and the vertical axis represents mutual information estimation values. The blue and orange lines represent the methods without and with unlabeled data, respectively.}
	\label{fig:mi}
\end{figure}

\subsection{Effect of unlabeled data}
\label{ssec:effect-num-unlabeled-data}
1) \textbf{Mutual information estimation:}
The most direct reason for using unlabeled data is to estimate mutual information more accurately \cite{CLUB}.
We plot the line chart of mutual information estimation values during the training process, as shown in Figure \ref{fig:mi}.
The mutual information estimation values refer to the sum of mutual information between language, visual, audio, and shared representations pair-wise in each epoch, averaged over batches.
From the figure, we can observe that without unlabeled data, the estimated mutual information converges rapidly (converging to approximately 0.4) but then exhibits intense oscillations. 
On the other hand, with unlabeled data, the mutual information estimation continuously decreases, approaching convergence at a smaller value 0.2, and the overall trend remains smooth.
Therefore, the addition of unlabeled data can lead to more accurate mutual information estimation and more stable training process.

2) \textbf{Model performance:}
The performance improvement resulting from the utilization of unlabeled data may be attributed to better estimating mutual information, thereby eliminating redundancy in multimodal joint representations. It may also be due to assisting in characterizing the structure of multimodal data by reconstruction, preventing model overfitting.
To explore this, we conducted experiments on the CMU-MOSI dataset, and the results are shown in Table \ref{tab:ablation-unlabeled-mosi}.
In the table, \textcircled{\raisebox{-0.8pt}{1}} refers to MIRD$^{2}$, which uses unlabeled data to estimate mutual information and reconstruct the original input; \textcircled{\raisebox{-0.8pt}{2}} indicates using unlabeled data to estimate mutual information without reconstructing the original input; \textcircled{\raisebox{-0.8pt}{3}} represents using unlabeled data to reconstruct the original input without estimating mutual information; \textcircled{\raisebox{-0.8pt}{4}} refers to MIRD$^{1}$, which does not use any unlabeled data.
Through the comparison of \textcircled{\raisebox{-0.8pt}{1}} with \textcircled{\raisebox{-0.8pt}{2}} and \textcircled{\raisebox{-0.8pt}{3}}, as well as \textcircled{\raisebox{-0.8pt}{4}} with \textcircled{\raisebox{-0.8pt}{2}} and \textcircled{\raisebox{-0.8pt}{3}}, we observe that the improvement in model performance is not only attributed to unlabeled data assisting in estimating mutual information but also to its contribution in reconstructing the original input and capturing the underlying structure of multimodal data.

\begin{table}[h]
	\centering
	\caption{\centering Comparison experiments on CMU-MOSI datasets.}
	\label{tab:ablation-unlabeled-mosi}
	\begin{tabular}{c|*4{c}}
		\toprule[1pt]
		\textbf{Method} &\textbf{Acc} $\uparrow$ &\textbf{F1} $\uparrow$ &\textbf{MAE} $\downarrow$ &\textbf{Corr} $\uparrow$ \\
		\hline
		\rule{0pt}{1.0em}\textcircled{\raisebox{-0.8pt}{1}} &86.23\% &86.16\% &0.721 &0.804 \\
		\rule{0pt}{1.0em}\textcircled{\raisebox{-0.8pt}{2}} &86.05\% &86.04\% &0.727 &0.800 \\
		\rule{0pt}{1.0em}\textcircled{\raisebox{-0.8pt}{3}} &85.98\% &85.92\% &0.723 &0.801 \\
		\rule{0pt}{1.0em}\textcircled{\raisebox{-0.8pt}{4}} &85.82\% &85.74\% &0.723 &0.799 \\
		\bottomrule[1pt]
	\end{tabular}
\end{table}

\begin{figure}[h]
	\centering
	\includegraphics[trim = 0mm 0mm 0mm 2mm, clip=true, width=0.8\linewidth]{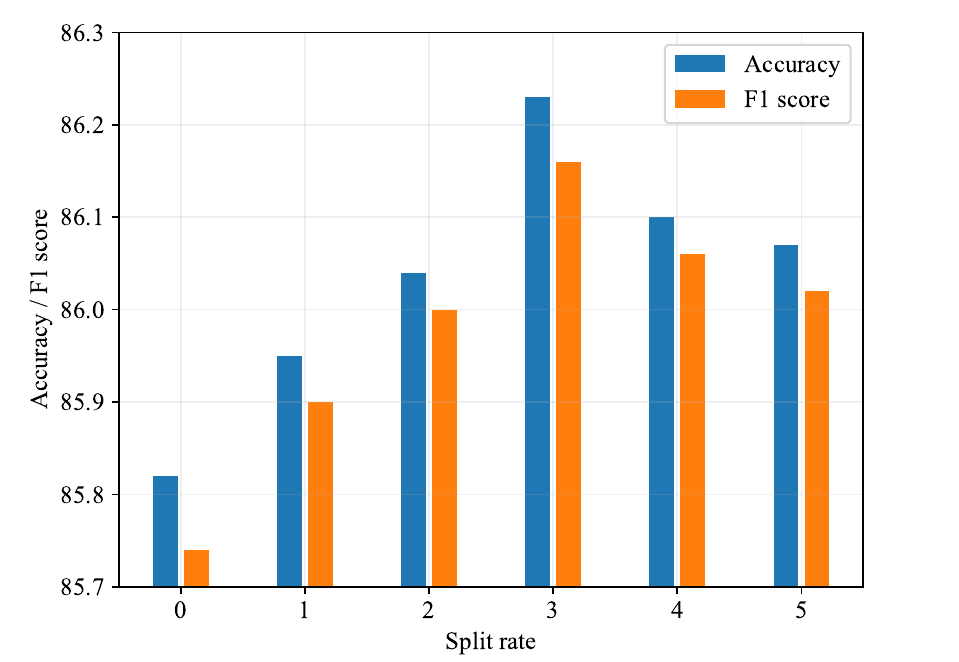}
	\caption{Bar chart illustrating the impact of unlabeled data quantity on model performance on CMU-MOSI dataset. The split rate represents the ratio of unlabeled data to labeled data (the split rate 0 implies the absence of unlabeled data). The blue and orange bars represent accuracy and F1 score, respectively.}
	\label{fig:performance}
\end{figure}

In addition, to investigate the impact of the amount of unlabeled data on model performance, we present the bar chart in Figure \ref{fig:performance}.
From the figure, we observe that as the amount of unlabeled data increases, the model performance improves. 
The model achieves its best performance (86.23\% on accuracy, 86.16\% on F1 score) when the split rate is equal to 3. 
However, further increasing the amount of unlabeled data results in a gradual decline in performance.
This is because (1) the model may have already learned sufficient information from the unlabeled data, and further increasing the data does not significantly improve the performance.
(2) given the limited capacity of the model, extensive unlabeled data might hinder the model from effectively utilizing all the available data. 
Instead, it could introduce excessive noise, potentially leading to a decline in model performance.

\begin{figure}[t]
	\centering
	\includegraphics[trim = 0mm 0mm 0mm 2mm, clip=true, width=0.8\linewidth]{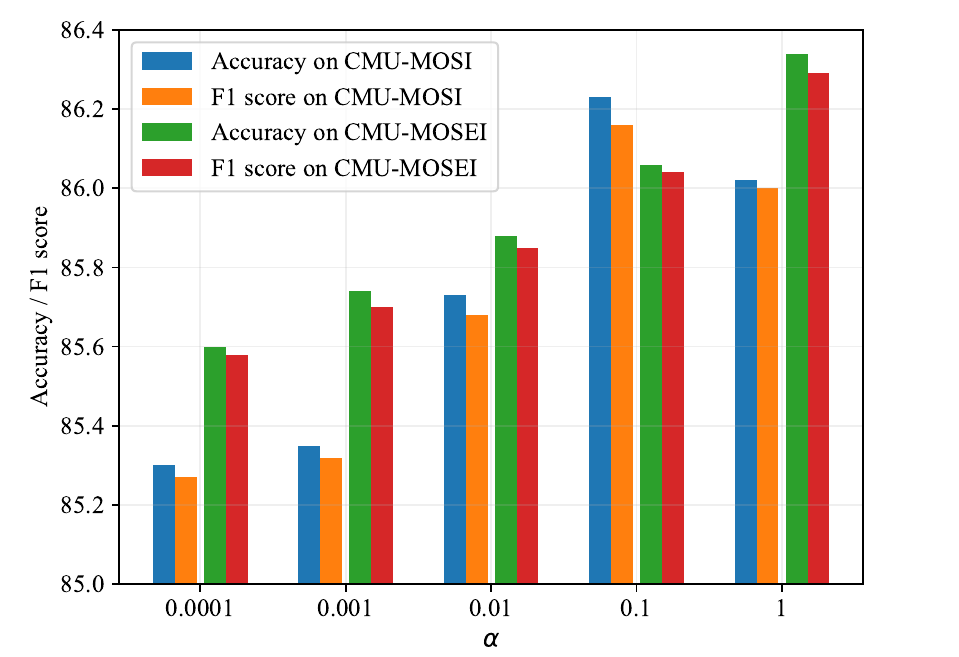}
	\caption{Bar chart of the model performance under different hyper-parameter $\alpha$ values on CMU-MOSI and CMU-MOSEI datasets. The blue and orange bars represent accuracy and F1 score on CMU-MOSI dataset. The green and red bars represent accuracy and F1 score on CMU-MOSEI dataset.}
	\label{fig:alpha}
\end{figure}

\subsection{Effect of loss weights}
\label{ssec:loss-weight}
In order to select the optimal loss weight $\alpha$, we compare the performance of the model under different loss weights on CMU-MOSI and CMU-MOSEI dataset, as shown in Figure \ref{fig:alpha}.
From the figure, we can observe that on the CMU-MOSI dataset, the best performance is achieved when $\alpha$ equals to 0.1, while on the CMU-MOSEI dataset, the best performance is achieved when $\alpha$ equals to 1.
This indicates that the weight of mutual information minimization should neither be too large nor too small. 
If it is too large, it can make the model difficult to learn, while if it is too small, it may not effectively serve as a regularization term.
In addition, we find that when $\alpha$ equals to 10, the performance of the model was significantly poor, with 58.23\% accuary and 58.20\% F1 score on CMU-MOSI dataset, 58.08\% accuracy and 58.02\% F1 score on CMU-MOSEI dataset (for aesthetic reasons, we did not depict the performance when $\alpha$ equals to 10 in the figure).
The reason is that the model prioritize disentangling correlations between latent representations, thereby neglecting the primary task of sentiment prediction.

\subsection{Case study}
\label{ssec:case-study}
\begin{figure*}[t]
	\centering
	\includegraphics[trim = 0mm 0mm 0mm 2mm, clip=true, width=\textwidth]{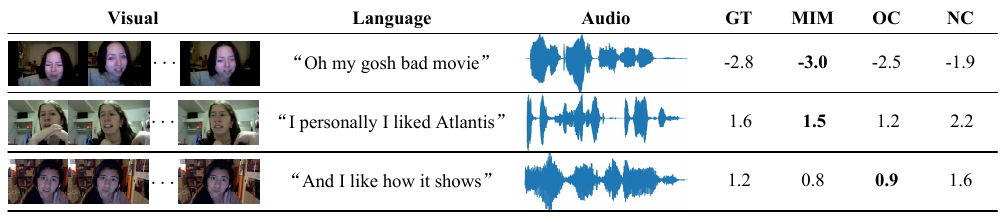}
	\caption{Examples from the CMU-MOSI dataset. For each example, we show the Ground Truth (GT) and prediction output of the model with Mutual Information Minimization (MIM), Orthogonal Constraint (OC), No Constraints (NC).}
	\label{fig:case-study}
\end{figure*}

To visually validate the reliability of our model, we present three examples shown in Figure \ref{fig:case-study}.
The examples are from the first three speaker videos of the test set of the CMU-MOSI dataset.
From the figure, we can observe that the MIM-based approach exhibits the best sentiment prediction performance on the first two examples, and it is also comparable to the OC-based approach on the last example. 
This validates the generalizability of our proposed method in sentiment prediction tasks. 
Furthermore, we note that the method incorporating latent representation constraints (MIM, OC) outperforms unconstrained method (NC) in terms of sentiment prediction ability, indicating the significance of representations disentanglement.

\section{Conclusion}
\label{sec:conclusion}
In this paper, we highlighted the importance of learning modality-agnostic and modality-specific representations on modeling unaligned multimodal language sequences. 
A novel representations disentanglement architecture is proposed to learn a single modality-agnostic representation.
The mutual information minimization constraints are further introduced along with our architecture to achieve superior disentanglement of representations, and avoiding information redundancy in multimodal joint representation.
In addition, we emphasized the significance of employing unlabeled data.
On one hand, it aids in estimating mutual information, and on the other hand, it helps to characterize the underlying structure of multimodal data, further mitigating the risk of model overfitting.
Our method significantly improves the baseline and outperforms the state-of-the-art models in experiments.

\bibliographystyle{elsarticle-num}
\bibliography{mybibfile}



\end{document}